\documentclass{article}
\usepackage[dblblindworkshop,final]{neurips_2026}
\workshoptitle{Interpretability as a Science}
\usepackage[utf8]{inputenc}
\usepackage[T1]{fontenc}
\usepackage{graphicx}
\usepackage{booktabs}
\usepackage{tabularx}
\usepackage{float}
\usepackage{microtype}
\usepackage{xcolor}
\usepackage{amsmath,amssymb}
\usepackage{hyperref}
\usepackage{url}
\hypersetup{hidelinks}

\usepackage{amsmath,amsfonts,bm}

\def\eqref#1{equation~\ref{#1}}

\def\1{\bm{1}}

\DeclareMathAlphabet{\mathsfit}{\encodingdefault}{\sfdefault}{m}{sl}
\SetMathAlphabet{\mathsfit}{bold}{\encodingdefault}{\sfdefault}{bx}{n}

\newcommand{\Accept}{\textsc{Accept}}
\newcommand{\Reject}{\textsc{Reject}}
\newcommand{\Abstain}{\textsc{Abstain}}
\newcolumntype{Y}{>{\raggedright\arraybackslash}X}

\title{Towards Computational Provenance:\\
Carrying Causal-State Evidence in Generated Text}

\author{
  Benjamin Belay \\
  Independent Researcher \\
  \texttt{benjamin.belay@hotmail.co.uk}
}

\begin{document}
\maketitle
\begin{abstract}

A language model's output does not by itself provide verifiable evidence about
the internal computation that produced it. We study \textit{computational provenance}: whether generated text can carry detectable evidence of which causally relevant
internal state occurred. We test a bounded form of this idea in two controlled
architectures: a modular feed-forward neural network and a transformer-based model. Both architectures are trained on the same arithmetic task with a mandatory pathway through
two discrete intermediate states, allowing different internal paths to produce the same answer.
We deliberately switch between these paths, authenticate the state actually used, and let that
verified state determine a subtle statistical pattern in the generated text that can later be detected.
The feed-forward and transformer systems each passed all 128 matched pairs in both their public and separately sealed protected end-to-end evaluations, with the detector recovering the signal associated with the authenticated internal state. The required causal computation also reproduced across five independently trained feed-forward models and three independently trained transformers. In a separate answer-only transformer experiment, our linear probes did not
recover a naturally learned intermediate state. These results provide a controlled proof of concept
that information about a verified, causally relevant internal state can be
preserved in generated text even when the answer is unchanged.

\end{abstract}
\section{Introduction}

Language-model-based AI systems increasingly produce answers, plans,
explanations, tool calls, and other generated records used in model evaluation,
process supervision, and auditing
\citep{lightman2023verify,bowman2022scalable}, and also play an important role
in approaches to scalable oversight
\citep{irving2018debate,burns2024weak}. These records often assume that a
generated artifact bears some meaningful relation to the computation that
produced it. Yet two executions can produce the same
answer and similarly plausible explanations while reaching that answer through
different internal computations \citep{jain2019attention,mcgrath2023hydra}: an apparently reasonable explanation does
not establish its own causal origin. Chain-of-thought, for example, is generated text rather than a direct
observation of the underlying computation, and may omit or rationalise
important influences or compress several operations into a simpler account
\citep{turpin2023unfaithful,lanham2023faithfulness}. Other interpretability methods inspect the model more directly: sparse
autoencoders extract interpretable features from internal activations
\citep{huben2024sparse}; Natural Language Autoencoders translate activations
into readable descriptions \citep{frasertaliente2026nla}; and J-space methods
identify internal representations that the model is likely to express in its output
\citep{gurnee2026verbalizable}. These methods can reveal
information represented inside the model, but do not by themselves establish
authenticated provenance of the particular causal computation that produced an
output. This
distinction becomes especially important if a model has hidden objectives or
behaves strategically, since its stated reasoning may not reveal the internal
factors that actually drove its behaviour \citep{hubinger2024sleeper,scheurer2023strategic}. 

As these systems are applied to
tasks that are increasingly difficult for human supervisors to evaluate
directly, oversight may require evidence not only that an answer appears
acceptable, but that the generated artifact remains connected to the
computation that produced it. We ask whether evidence of such differences in
internal computation can be preserved in the generated output.

This work asks whether that connection can be made verifiable. Rather than
attempting to reconstruct a model's complete internal reasoning or build a
general deception detector, we study a narrower question: can we identify a
causally relevant internal state, verify which state occurred, and make that
state determine a detectable signal in the model's generated output? We call
this \textit{computational provenance}. Crucially, we ask whether evidence of different internal states can still be
preserved across executions that differ internally but have the same prompt,
final answer, semantic content, and sampling randomness. The aim is not to
recover the model's full internal computation from its generated output, but to preserve
evidence about a causally relevant part of the computation that occurred. Such
a signal could complement interpretability and oversight by distinguishing
answer-equivalent executions that followed different internal paths but would
otherwise appear the same to an evaluator.

We study this question using two purpose-built models trained on the same
arithmetic task: a small modular feed-forward neural network and a
transformer-based model. In both constructions, every answer must pass through
two discrete intermediate states, $z_2$ and $z_3$. We run the same prompt
twice---once naturally and once after replacing $z_2$---so that the executions
return the same final answer through different internal paths. After verifying
which state each execution used, that state determines a statistical pattern
during text generation, and a detector tests which of the possible state
patterns is present in the resulting output. In both architectures, we use the same process to authenticate the state,
carry its statistical pattern into the text, and detect that pattern in the
output.
Neither construction is intended to reproduce the scale or open-ended
behaviour of a modern language model; they provide controlled settings in which
an internal path can be changed and its downstream effects measured directly.
Figure~\ref{fig:system-architecture} summarises this pipeline.


\begingroup
\setlength{\intextsep}{6pt}
\setlength{\abovecaptionskip}{3pt}
\setlength{\belowcaptionskip}{0pt}

\begin{figure}[H]
  \centering
  \includegraphics[
    width=\linewidth,
    trim=0 220bp 0 130bp,
    clip
  ]{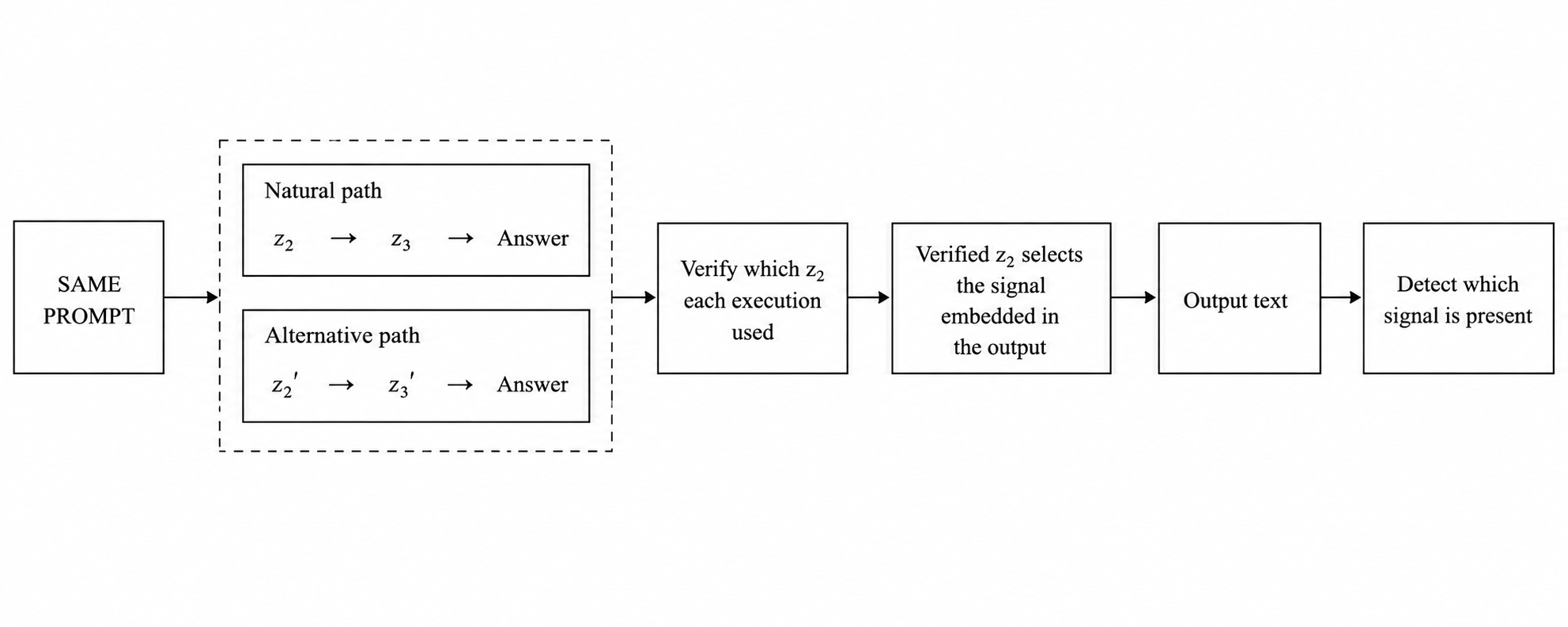}
  \caption{Overview of the experimental construction, implemented with both
  feed-forward and transformer state modules. Two executions produce the same
  answer through different internal paths. The verified intermediate state
  selects a statistical signal during text generation, and detecting that
  signal provides evidence of which state occurred.}
  \label{fig:system-architecture}
\end{figure}

\endgroup

In the feed-forward construction, the fixed training procedure reproduced the
required causal pathway across five fresh models. A separately trained model
then achieved 128/128 on both public and protected end-to-end evaluations. We
next replaced the feed-forward state modules with two transformer encoders
while keeping the discrete pathway and provenance mechanism fixed. The
transformer pathway reproduced across three fresh models, and a separately
reserved transformer likewise achieved 128/128 publicly and 128/128 on a
prospectively sealed protected set, without recalibrating the text signal or
detector. We also tested a \textit{natural-state} setting in which three additional transformers were
trained only to produce the final answer, without supervision for $z_2$ or
$z_3$. Although all three learned the task perfectly, frozen linear probes did not
recover a qualifying intermediate state in the designated development model,
so causal intervention and provenance testing were not attempted in the
natural-state setting. Together, these
results establish a controlled proof of concept for \textit{computational
provenance}: a verified, causally relevant internal
state can determine a detectable pattern in generated text, even when the final
answer is unchanged. Finding suitable internal states in larger language models
remains an open problem.
\section{Related work}

Existing work provides ways to study internal model states, verify aspects of
model execution, and place detectable signals in generated text. What remains
less explored is whether the generated output can preserve evidence about the
model's own internal computation.

One way to make a model's internal computation easier to study is to require it to pass through intermediate states with predefined meanings before producing its final answer. Concept
bottleneck models use this structure so that researchers can inspect these
states, intervene on them, and measure how they affect the model's behaviour
\citep{koh2020concept,shin2023closer}. Causal-abstraction methods address a
related question: whether these internal states actually have the causal role
we think they do. If a state is changed, does the model's later computation
change in the corresponding expected way?
\citep{geiger2021causal,geiger2022inducing}. Together, these approaches provide
ways to define meaningful internal states and test whether the model actually
uses them. We build on this idea and ask a further question: once a causally
relevant state has been identified, can evidence of which state occurred be
carried beyond the internal computation into the model's generated output?

A separate line of work verifies the origin or execution of model outputs.
SafetyNets provides mathematical proofs that outsourced neural-network
inference was computed correctly, while Slalom uses trusted hardware to verify
neural-network operations delegated to an untrusted processor
\citep{ghodsi2017safetynets,tramer2019slalom}. SVIP tests whether a remote
provider used the claimed language model, using processed hidden
representations as model-specific evidence \citep{sun2025svip}. C2PA binds
signed claims about the origin and editing history of digital content, while
AEX binds an API request to its response and subsequent transformations
\citep{c2pa2025,guan2026aex}. These approaches verify the model, execution, or
history associated with an output, but they do not distinguish executions that
use the same model and produce the same answer while following different
causally relevant internal paths. We address this complementary question by
verifying which internal state occurred, showing that it affected the later
computation, and preserving a detectable signal of that state in the generated
text.

Text watermarking places detectable statistical signals in generated language
by slightly biasing token choices according to a secret key
\citep{kirchenbauer2023watermark,kuditipudi2024robust}. SynthID-Text uses this
general approach to identify model-generated text
\citep{dathathri2024scalable}. More recent methods condition watermarking on
model-associated information: ReasonMark uses written reasoning
\citep{liu2026reasonmark}, SAEMark uses learned model features
\citep{yu2025saemark}, SLAM manipulates model features to induce a chosen
watermark \citep{harelcanada2026slam}, and BiCoT introduces an ownership signal
during reasoning \citep{lu2026bicot}. Our aim is different: rather than using
model-associated information to support generation identification or
ownership, we use a verified causally relevant state to determine the signal
itself, so that the output preserves evidence of which state occurred even when
the final answer is unchanged.
\section{Construction}

To study how a model's internal computation can be linked to generated text, we
implement the same controlled arithmetic pathway in two model architectures: a
modular feed-forward network and a transformer-based model. In each case,
interventions test whether the intermediate states affect later computation;
cryptographic records authenticate which states were used; and the verified
state determines a statistical pattern carried into generated text, which a
detector then tests for in the final output.

\subsection{A mandatory discrete-state pathway}

We use a small arithmetic task that exploits modular arithmetic to allow two
different internal state paths to produce the same final answer. The task takes an input prompt of four numbers,
$x=(a,b,c,d)$, each between 0 and 15, with the model trained across many such
prompts, of different values of $a$, $b$, $c$, and $d$. For each prompt, the
target computation is
\begin{equation}
z_1=(a+b)\bmod16,\qquad
z_2=(z_1+c)\bmod16,\qquad
z_3=(5z_2+d)\bmod16.
\end{equation}

The models then produce the final answer
\[
y=z_3\bmod8,
\]
giving the learned pathway
\[
\text{prompt }x \longrightarrow z_2 \longrightarrow z_3 \longrightarrow y.
\]

Both architectures are built so that the computation must pass through two
explicit, discrete states, $z_2$ and $z_3$. The module that produces $z_3$
receives only $z_2$ and $d$, not the earlier inputs $a$, $b$, or $c$, and the
answer module receives only $z_3$. There is therefore no route around either
state. The auxiliary value $z_1$ is used only to define $z_2$ and is not itself
a model state.

The final modulo-8 operation is crucial to the experiment because it allows two different internal paths to produce exactly the same answer. Values of $z_3$ that differ by 8 map to the same value of $y$; for example:
\[
5\bmod8 = 13\bmod8 = 5.
\]

We therefore use values 0--15 and modulo 16 so that every internal state has a
corresponding state 8 values away. To create the matched executions used in the
experiment, we run each prompt twice. First, the model runs normally
and produces its natural value of $z_2$. We then run the same prompt again,
but replace that value with
\begin{equation}
z_2'=(z_2+8)\bmod16.
\end{equation}

The calculation of $z_3$ is designed so that this change in $z_2$ carries
forward to the next internal state. In particular, adding 8 to $z_2$ also adds
8 to $z_3$ modulo 16:
\begin{align}
z_3'
  &= \bigl(5z_2' + d\bigr)\bmod16 \\
  &= (z_3+8)\bmod16.
\end{align}
Thus the intervention changes both $z_2$ and $z_3$, while the final modulo-8
step removes that difference:
\[
z_3'\bmod8=z_3\bmod8.
\]
The two executions therefore follow different internal paths but still produce
the same observed answer.

For example, consider the prompt $x=(1,1,2,1)$. Its natural execution gives
\[
z_1=2,\qquad z_2=4,\qquad z_3=5,\qquad y=5.
\]
The alias intervention replaces $z_2=4$ with $z_2'=12$. The model's downstream
transition then gives $z_3'=13$, but the answer remains 5:
\[
\begin{aligned}
\text{natural:}\quad
  z_2=4  &\longrightarrow z_3=5
          \longrightarrow 5\bmod8=5,\\
\text{alias-intervened:}\quad
  z_2'=12 &\longrightarrow z_3'=13
           \longrightarrow 13\bmod8=5.
\end{aligned}
\]
The two executions therefore receive the same prompt and return the same answer,
but pass through different $z_2\rightarrow z_3$ states.

The arithmetic specifies the intended relationship between the states, but the
provenance claim also requires $z_2$ to affect the model's later computation.
We therefore intervene on $z_2$ and test whether $z_3$ changes as predicted.
The feed-forward construction uses separate multilayer modules for the two
states, while the transformer construction replaces them with separate
transformer encoders. In both cases, only the selected 16-way $z_2$ state passes to
the second stage, and only the discrete $z_3$ state is used to produce the final answer. This
preserves the same causal interface across both architectures.

\subsection{Recording and verifying the internal state}
\label{sec:state-verification}

To distinguish which internal path the model actually took, we record and verify
the intermediate states used during each execution. Trusted instrumentation
observes $z_2$ and $z_3$ at the points where they are used in the model's
computation and records those events directly.

Each recorded state is stored in a small cryptographically protected record,
which we call a \emph{receipt}. Each receipt contains the recorded state and a message authentication code computed with a secret key, allowing the verifier to detect alteration or fabrication.

We use two kinds of state evidence. An \emph{exact receipt} identifies the
particular intermediate-state event from one execution. An \emph{abstract
receipt} records the corresponding state value, such as $z_2=4$, which is the
identity used to select the later state-specific statistical signal. Thus
different executions that genuinely use the same $z_2$ value select the same
signal, while executions that use different $z_2$ values select different
signals. We also record the $z_2\rightarrow z_3\rightarrow y$ sequence so that
the verifier can check which internal path was taken and whether it is
consistent with the model's computation.

Only after these checks succeed is the authenticated value of $z_2$ allowed
to determine the statistical signal used during text generation. The resulting signal is therefore tied to a state that was actually observed
during the model's computation, rather than to a value chosen afterwards.

We assume that the instrumentation, verifier, and cryptographic keys have not
been compromised; this trust assumption and its implications are discussed in
Section~\ref{sec:limitations}.

\subsection{Carrying the internal state into generated text}

After the receipts have established which value of $z_2$ the model actually
used, we ask whether evidence of that internal state can be causally carried
into the model's generated text. Where the arithmetic task provides the controlled
internal computation, the text provides an observable output in which a detectable trace of that computation is deliberately introduced. We therefore generate text with
fixed meaning but allow several words in a fixed-length output sentence to vary between equivalent alternatives.
This allows different internal states to
leave different statistical patterns in the wording without changing what the
text reports.

For the output, we construct a short textual report accompanying the
computation. Its numerical content is fixed; only some of the surrounding
words are allowed to vary. We introduce six variable word positions in each sentence so
that the verified internal state can influence several small wording choices
rather than being represented by a single word. Each sentence follows the same
template:
\[
[\text{Group 1}]\ \text{the}\ [\text{Group 2}]\ [\text{Group 3}]\
[\text{Group 4}]\ [\text{Group 5}]\ [\text{field}]\ \text{is}\
[\text{value}]\ [\text{Group 6}].
\]

Each group contains eight permitted alternatives. For example, Group 1 may
contain
\[
\{\text{Here, Now, Presently, Directly, Briefly, Simply, Accordingly,
Formally}\}.
\]

For each verified value of $z_2$, four of the eight alternatives in every
group are designated as \emph{favoured} and the other four as
\emph{unfavoured}. The favoured alternatives are made slightly more likely to
be chosen. Different values of $z_2$ favour different subsets of words, rather than $z_2$ directly selecting a particular word.

This is how the two
answer-equivalent internal paths can produce different wording patterns even
though they report the same result. For example, suppose the natural execution uses $z_2=4$ and the alternative
execution uses $z_2'=12$, while both produce the same final answer $y=5$.
Possible outputs include:
\[
\begin{aligned}
z_2=4:\quad
&\text{``}\underline{\text{Here}}\ \text{the}\
 \underline{\text{calculation}}\
 \underline{\text{explicitly}}\
 \underline{\text{shows}}\
 \underline{\text{that}}\
 \text{answer is 5}\
 \underline{\text{therefore}}\text{.''}\\
z_2'=12:\quad
&\text{``}\underline{\text{Now}}\ \text{the}\
 \underline{\text{derivation}}\
 \underline{\text{precisely}}\
 \underline{\text{reports}}\
 \underline{\text{namely}}\
 \text{answer is 5}\
 \underline{\text{thus}}\text{.''}
\end{aligned}
\]

The complete text contains four short statements labelled $z_1$, $z_2$, $z_3$,
and $y$, whose numerical content is fixed across matched executions; the
receipts separately record the authenticated states. Each statement has six variable word positions, giving 24 word choices in total. The small preference for favoured words can therefore accumulate across the text into a statistical pattern associated with the verified value of $z_2$.

To isolate the effect of the internal state, the natural and alternative
executions use the same random draw at each corresponding word position. Any
resulting difference in wording therefore comes from the state-dependent
preferences rather than different sampling randomness.

The detector then compares the 24 observed word choices with the patterns
associated with each of the 16 possible values of $z_2$. It uses a fixed
statistical scoring rule, rather than a trained classifier, and gives higher
scores when more of the observed words match those favoured by a candidate
state. The detector accepts a state only when its score exceeds a threshold
fixed on separate calibration data and is higher than the scores for all other
candidate states. Detection therefore identifies which state-specific wording
pattern is present while the meaning of the generated text remains unchanged.

Details of the scoring rule, normalisation, sampling procedure, word groups and values, and calibration
are given in Appendix~B.2.
\section{Experimental design}

We evaluate three parts of the system: whether the model uses the intermediate
state causally, whether that state can be verified, and whether the associated
signal can be recovered from the generated text. We first test each part
separately, then evaluate the complete pipeline on public and protected data.

\subsection{Models and causal tests}

The feed-forward model has 12,648 parameters. A 128-unit hidden layer produces
the 16-way state $z_2$, a 64-unit hidden layer produces the 16-way state $z_3$,
and a final linear layer maps $z_3$ to one of eight answers. Five independently
initialised models are trained for 5,000 steps and evaluated at the fixed
step-5,000 checkpoint. The engineered transformer has 546,088 parameters and uses two separate
two-layer transformer encoders. Each encoder has width 128, four attention
heads, and a 256-unit feed-forward block. The first encoder receives tokens for
$a$, $b$, and $c$ and produces the 16-way state $z_2$. The second receives
only that state and $d$ and produces the state $z_3$; the final layer
receives only $z_3$. Three independently initialised transformers are trained
for 12,000 steps and evaluated at the fixed step-12,000 checkpoint. The models
are supervised on $z_2$, $z_3$, and the answer, as in the feed-forward
construction.

We first measure whether each model computes the correct $z_2$, $z_3$, and answer on 4,096 held-out inputs, and then intervene on its internal states. The main intervention replaces $z_2$ with $(z_2 + 8) \bmod 16$, which should change $z_3$ while preserving the answer, and we also test an answer-changing intervention together with same-state, wrong-state, sham, and direct-$z_3$ controls. These tests determine whether later computation responds to the value of the consumed state rather than simply to the act of intervention. Additional training details are given in Appendix~A.5.

\subsection{State verification and signal detection}

Trusted instrumentation records the $z_2$ and $z_3$ values used during an
execution, and the resulting records are authenticated using keyed hashes
(HMACs)\footnote{An HMAC is a short cryptographic tag calculated from a record
and a secret key. A verifier with the same key can check whether the record has
been changed; it does not hide the record's contents.}. This allows the verifier
to check which states occurred, the execution they belong to, and the order in
which they were used, and we test this mechanism
using valid records together with altered, replayed, reordered, mismatched, and
missing records.

Once $z_2$ has been verified, its value determines which statistical signal is
used during text generation. 
Each execution produces eight reports. The fixed detector combines the evidence
across them and scores the patterns associated with all 16 possible values of
$z_2$. It identifies a state only when its score exceeds a threshold set using
separate calibration data and is higher than every competing score.

Both architectures use the same 16 state-specific patterns, number of reports,
scoring rule, and thresholds. Only the model and checkpoint identities recorded in the receipts change; the text signal and detector are not recalibrated.

\subsection{End-to-end evaluation}

The main comparison uses a natural and an alternative execution of the same
prompt:
\[
z_2\rightarrow z_3\rightarrow y,
\qquad
z_2'\rightarrow z_3'\rightarrow y.
\]
The two executions follow different internal paths while keeping the prompt,
final answer, semantic content, generation settings, and position-indexed
sampling draws fixed. We then test whether changing the verified value of
$z_2$ produces the corresponding change in the statistical signal detected in
the generated text. A pair is counted as successful only when both executions follow the expected
computation, their receipts verify correctly, and the detector identifies the
signal associated with the verified state. We also include controls in which no
signal is added, a signal from another state is used, or the receipt evidence is
invalid.

For each architecture, we train one further model using its fixed training
procedure and reserve a protected set of 128 matched pairs before training. The model is first required to pass the public computation and causal
tests, followed by the public end-to-end evaluation on 128 matched pairs. It is
then evaluated once on the sealed protected set, with the model, generator,
detector, thresholds, and receipt rules kept unchanged throughout.

\subsection{Answer-only state localisation}

Separately from the engineered transformer, we ask whether a similar internal state emerges naturally in a
transformer that is trained only to produce the correct answer, without being
explicitly taught the intermediate states $z_2$ and $z_3$. We
train three four-layer, 540,808-parameter transformers with width 128, four
attention heads, and 256-unit feed-forward blocks. After confirming task performance, we extract residual-stream activations from
20 fixed combinations of layer and token position in one designated
development model and train frozen linear probes on those activations. We test
whether they encode the full 16-way $z_2$ state, the part of $z_2$ needed
to determine the final answer, and the remaining distinction between
answer-equivalent states such as $z_2$ and $(z_2+8)\bmod16$. The other two
models are used for replication only if the development model satisfies the
predefined probe-performance criteria.
\section{Results}

Both engineered architectures passed their final evaluations. The required
causal pathway reproduced across all five feed-forward models and all three
transformer models. A separately trained model of each architecture then passed
its public computation and causal tests before achieving 128/128 on both its
public and protected end-to-end evaluations. Table~\ref{tab:final-results}
summarises the main results.

\begin{table}[t]
\caption{Causal robustness and end-to-end provenance in the two engineered
architectures. The robustness and provenance rows use separate models.}
\label{tab:final-results}
\centering
\small
\begin{tabularx}{\linewidth}{@{}lXr@{}}
\toprule
Architecture & Evaluation & Result \\
\midrule
Feed-forward
  & Causal robustness across independent models
  & 5/5 \\
Transformer
  & Causal robustness across independent models
  & 3/3 \\
Feed-forward
  & Public / protected end-to-end provenance
  & 128/128 / 128/128 \\
Transformer
  & Public / protected end-to-end provenance
  & 128/128 / 128/128 \\
Answer-only transformer
  & Task competence / qualifying probed sites
  & 3/3 / 0/20 \\
\bottomrule
\end{tabularx}
\end{table}

\subsection{Causal pathway and component tests}

Across both architectures, the answer-preserving intervention
changed $z_3$ without changing the final answer, whereas the answer-changing
intervention produced the predicted different answer. Same-state, wrong-state,
sham, and direct-$z_3$ controls also behaved as expected. A separately reserved
model of each architecture then passed all 4,096 public computation cases and
all 28,672 causal cases before entering the provenance evaluation.

The receipt and text-signal components were validated before integration.
Valid records authenticated the consumed states and their order, while altered,
replayed, reordered, mismatched, and missing records produced their expected
outcomes. The detector operating point was fixed at
$T=4.041451884327381$ and $M=0$ before the end-to-end evaluations and was
unchanged for both architectures. Sample sizes and outcomes for each component are reported in
Appendix Table~\ref{tab:stage_denominators}.

\subsection{End-to-end provenance}


The designated feed-forward model achieved 128/128 on its public evaluation
and 128/128 on its separately sealed protected evaluation. In every successful
pair, the natural and alternative executions followed different
$z_2\rightarrow z_3$ paths while preserving the prompt, final answer, semantic
content, and position-indexed sampling draws. The detected signal changed with
the authenticated value of $z_2$.

The same provenance mechanism was then transferred to the engineered
transformer. Its separately designated model also achieved 128/128 publicly
and 128/128 on a protected set sealed before training. The transformer used the
same 16 authorities, text generator, detector, output count, and calibrated
thresholds as the feed-forward construction. Across the transformer's public and protected evaluations, there were no wrong-state confusions, no detections in the
no-signal controls, and no failed receipt or registered integrity checks.

Appendix~\ref{app:example} shows a paired output example. Both executions
produce answer 6, while the authenticated state changes from $z_2=2$ to
$z_2=10$ and the detected wording pattern changes accordingly.

\subsection{Answer-only state localisation}


The three fresh answer-only transformers each achieved 100\% overall,
minimum-class, and minimum-template accuracy. In the designated development
model, however, no frozen linear probe met the recovery criteria at any of the
20 predefined token-level residual sites. The best accuracy was 23.80\% for
the full 16-way $z_2$ value and 45.54\% for $z_2\bmod8$.

Because no site qualified, we did not proceed to causal intervention, examine
the two reserved models, or attach the provenance mechanism. This is a bounded
negative result for the tested model, locations, and linear probe family. It
does not show that the relevant information is absent or that natural
computational provenance cannot exist.
\section{Discussion}

The experiments show that computational provenance can operate in two
controlled learned architectures. In both a modular feed-forward neural network
and a transformer-based model, different internal paths could produce the same
answer while leaving different detectable patterns in the generated text.

\subsection{What the result establishes}

The causal pathway reproduced across five feed-forward models and three
transformers, and separately designated models of both architectures passed all
public and protected end-to-end evaluations. The transformer also reused the
same calibrated signal and detector, showing that the provenance mechanism was
not specific to the original feed-forward construction.

The result depends on combining three forms of evidence. Intervention tests
show that the recorded state participates in the computation rather than acting
as an unrelated label; authenticated receipts establish which state was
actually observed during a particular execution; and the statistical signal
preserves evidence of that verified state in the generated output. A receipt
or watermark alone would not provide this connection between causal internal
computation and observable text.

This suggests the possibility of a new verification framework for larger models. Methods such
as sparse autoencoders and Natural Language Autoencoders can identify or
describe candidate features in model activations, but computational provenance
adds a further question: did a particular state actually affect the computation
that produced this output? Candidate states could first be tested through causal intervention, then authenticated when they occur, and finally linked to a detectable signal carried by an observable part of the model's behaviour, such as an intermediate reasoning trace or final output. The detector would therefore
check evidence associated with verified internal computation rather than rely
only on the model's own explanation of its reasoning.

Such a system could complement interpretability and scalable oversight,
particularly when outputs are difficult for human supervisors to evaluate
directly. It would not recover a model's complete reasoning, but could provide
verifiable evidence that selected causally relevant parts of its internal
computation were connected to an observable output. Extending this from the
explicit states studied here to activation-level states in pretrained language
models is therefore an important next step.

\subsection{Scope and limitations}
\label{sec:limitations}

The present result remains a proof of concept in a finite arithmetic task with
an explicitly constructed discrete pathway. Both architectures are trained to produce $z_2$ and
$z_3$, and their structure forces the answer through those states. Although one implementation uses
transformer encoders, it is a small purpose-built transformer rather than a
pretrained language model. The
experiments therefore establish transfer across model architectures, not the
natural emergence of provenance-ready states in larger language models.

The text-generation setting is also deliberately constrained. Each report has
fixed semantic content and a limited vocabulary of interchangeable words, which
allows the statistical pattern to vary while meaning is held constant. Whether
the same approach remains effective for unrestricted generation, long-form
text, or outputs that are later edited or paraphrased remains to be tested.

The verification system assumes that the instrumentation, verifier, and secret
HMAC keys remain trustworthy. The receipt mechanism can detect the alterations
tested here, but it cannot protect against compromise of the components that
create or verify those records. The statistical signal should therefore not be interpreted as a trust-free or output-only proof that a computation occurred. Where a separate record can be stored, an authenticated sidecar containing
the state would provide a simpler and stronger integrity record, while the
text signal serves a different purpose by carrying evidence of that state
within the generated output itself, allowing the signal to remain detectable when the text is shared beyond its original execution environment, even without the associated execution records.

\subsection{Toward provenance in pretrained language models}

The answer-only transformer experiment illustrates an important boundary of
the present construction in a \textit{natural-state} setting. All three models solved the observable task
perfectly, but frozen linear probes did not recover a qualifying $z_2$
representation at the 20 predefined token-level locations in the designated
development model. High answer accuracy therefore did not automatically provide
a clear internal state that could be authenticated and carried into the output.
This result is deliberately narrow: the relevant information may be distributed,
represented nonlinearly, or organised differently from the predefined state. 

The next step is therefore not to assume that pretrained language models will naturally expose the same discrete states used here, but to identify or construct provenance-ready states within their activations. This may require learning candidate activation states through broader site searches and probes, or learning state representations that can be inserted and observed downstream, validating their causal role through intervention, and testing whether they support the same provenance mechanism without a purpose-built discrete pathway in the original model. Further work must also extend the output signal beyond constrained reports to less structured generation, including long-form text, paraphrasing, editing, and different decoding strategies.
\section{Conclusion}


We demonstrated computational provenance in two controlled model
architectures. In both a modular feed-forward network and an engineered
transformer, we changed the internal path while keeping the prompt, final
answer, semantic content, and sampling randomness fixed. Authenticated records
established which state was used, and generated text carried the corresponding
statistical pattern. The required causal pathway reproduced across five
feed-forward models and three transformers, while a separately designated model
of each architecture achieved 128/128 on both public and separately sealed
protected end-to-end evaluations. The transformer reused the same signal and
detector, showing that the mechanism transfers across architectures. The result
remains a controlled proof of concept with an explicitly constructed internal
pathway and constrained text generation, but it establishes a complete link
from causal internal computation to authenticated evidence and a detectable
signal in generated text. Extending the
approach to larger language models will require identifying or constructing
states within their activations that affect later computation. Computational
provenance could then support scalable oversight by providing independently
verifiable evidence about how an output was produced, rather than relying only
on the answer or the model's own explanation.

\bibliography{references}
\bibliographystyle{iclr2027_conference}

\appendix
\raggedbottom

\section{Additional experimental details}

\subsection{Evaluation details}

\begin{table}[H]
\caption{Sample sizes and outcomes for the principal evaluation stages.}
\label{tab:stage_denominators}
\centering
\small
\begin{tabularx}{\linewidth}{@{}lYY@{}}
\toprule
Stage & Sample size & Outcome \\
\midrule
Selected feed-forward causal confirmation
  & 512 examples; 3,584 computation, intervention, and control cases
  & all seven families passed \\

Five-model feed-forward computation study
  & 5 models; 4,096 held-out inputs per model
  & 5/5 models passed \\

Five-model feed-forward causal study
  & 28,672 intervention and control cases per model
  & 5/5 models passed \\

Receipt confirmation
  & 512 packages; 2,560 receipts; 9,216 attacks
  & all receipt and chain tests passed \\

Carrier calibration
  & 512 closures; 16,384 outputs including registered controls
  & fixed $T=4.041451884327381$ and $M=0$ \\

Generic carrier validation
  & 512 closures; 16,384 outputs including registered controls
  & met the predefined acceptance criteria \\

Designated feed-forward public qualification
  & 4,096 computation cases; 28,672 causal cases
  & all cases passed \\

Feed-forward public integration
  & 128 matched pairs
  & 128/128 passed \\

Feed-forward protected integration
  & 128 matched pairs
  & 128/128 passed \\

Engineered-transformer robustness study
  & 3 models; 4,096 computation cases and 28,672 causal cases per model
  & 3/3 models passed \\

Designated-transformer public qualification
  & 4,096 computation cases; 28,672 causal cases
  & all cases passed \\

Transformer public integration
  & 128 matched pairs
  & 128/128 passed \\

Transformer protected integration
  & 128 matched pairs
  & 128/128 passed \\
\bottomrule
\end{tabularx}
\end{table}

In the selected feed-forward model's causal confirmation, all 512 natural
executions produced the expected $z_2$, $z_3$, and answer. Answer-preserving, same-state, and sham interventions
preserved the expected answer, while answer-changing, wrong-state, and direct-$z_3$
interventions produced their predicted changes. After $z_2$ was replaced,
$z_3$ was still calculated by the neural network rather than assigned by the
controller.

The later feed-forward robustness study repeated the computation and
intervention tests on five freshly trained models. Each model was evaluated on 4,096
held-out inputs and 28,672 causal cases, and all five met the required
computation and causal criteria.

\subsection{End-to-end evaluation details}

Each 128-pair end-to-end evaluation contained 256 execution packages---128
natural and 128 alternative---and 1,280 receipts. For each package, the
generator produced eight state-conditioned reports and eight reports under
each of four control conditions, giving 40 outputs per package and 10,240
outputs in total. A matched pair passed only when both executions followed the
expected computation, their receipts verified, their generated content
remained correct, and the detector identified the signal associated with the
verified state while rejecting wrong-state and no-signal controls.

The prospective sequencing and outcomes of the feed-forward and transformer
protected evaluations are described in
Appendix~\ref{app:protected-evaluations}.

\subsection{Exact and abstract provenance}

Exact receipts, abstract receipts, and the statistical signal provide different
forms of evidence, summarised in Table~\ref{tab:provenance}.

\begin{table}[t]
\caption{Exact-event evidence is not abstract-state or carrier evidence.}
\label{tab:provenance}
\centering
\small
\begin{tabularx}{\linewidth}{@{}Yccc@{}}
\toprule
Candidate evidence & Exact & Abstract & Carrier claim \\
\midrule
Correct execution and state & \Accept & \Accept & compatible \\
Different execution, same state & \Reject & \Accept & compatible \\
Different registered state & \Reject & \Reject & incompatible \\
Required receipt missing & \Abstain & \Abstain & none \\
Present but invalid receipt & \Reject & \Reject & none \\
\bottomrule
\end{tabularx}
\end{table}

\subsection{Training and channel details}
\label{app:training-details}
The feed-forward models used AdamW with an initial learning rate of $10^{-3}$,
750 warm-up steps, weight decay $10^{-4}$, batch size 256, and gradient
clipping at 1.0. The learning-rate schedule was defined over a maximum of
10,000 steps. For the five-model feed-forward robustness study and the final
feed-forward evaluation, the evaluated checkpoint was fixed at step 5,000 for
every model; no checkpoint was selected separately for an individual seed.
The engineered transformers were trained separately for 12,000 steps and
evaluated only at the fixed step-12,000 checkpoint.

The constrained neural text generator, which supplies the base word
probabilities before the state-dependent bias is applied, used AdamW with
learning rate 0.003, weight decay $10^{-4}$, batch size 64, ten epochs, and
seed 817331. Its data contained 320 training, 96 calibration, and 96 untouched
channel-qualification semantic objects, each with eight variants.

\subsection{Engineered transformer construction}
\label{app:engineered-transformer}

The engineered transformer contains two separate two-layer transformer
encoders and 546,088 parameters. Each encoder uses width 128, four attention
heads, a 256-unit feed-forward block, GELU activations, pre-layer
normalisation, learned position embeddings, and no dropout.

The first encoder receives four tokens representing a learned classification
token and the fields $a$, $b$, and $c$. Its highest-scoring output is selected
as one of 16 discrete $z_2$ values. The second encoder receives a new
classification token, the selected $z_2$ value, and $d$, and similarly selects
one of 16 discrete $z_3$ values. The answer head receives only the selected
$z_3$ value. Parameters and residual streams are not shared between the two
transformer stages.

Three fixed robustness seeds were trained for 12,000 steps; all three passed
4,096 computation cases and 28,672 causal cases per model. A fourth,
prospectively designated transformer was then trained once under the same
recipe. It passed the same public computation and causal evaluations before
achieving 128/128 public and 128/128 protected end-to-end provenance results.
The protected population was selected and sealed before claim-bearing training.

The transformer used the previously calibrated carrier unchanged:
$T=4.041451884327381$, $M=0$, 16 candidate authorities, eight reports per
execution, 49 indexed draws per report, and 24 eligible word positions.
Receipt fields identifying the model and checkpoint were updated to bind the
new architecture, but the receipt semantics, carrier, detector, and attack
families were unchanged.

\subsection{Answer-only state localisation}
\label{app:answer-only}

To test whether the predefined intermediate state could be recovered without
direct intermediate supervision, we trained three 540,808-parameter,
four-layer transformers using final-answer cross-entropy only. Each model used width 128, four attention
heads, and 256-unit feed-forward blocks. All three achieved 100\% overall,
minimum answer-class, and minimum template accuracy on the frozen 4,096-row
held-out split.

Seed 772101 was designated for representation analysis. Linear logistic probes
were fitted on 6,144 examples and evaluated on a disjoint 3,072-example split.
We tested five token positions after each of four transformer layers, giving 20
predefined residual-stream locations. The probe targets included full $z_2$,
$z_2\bmod8$, the within-answer state distinction, $z_3$, and the answer.

The best full-$z_2$ accuracy was 23.80\%, and the best $z_2\bmod8$ accuracy
was 45.54\%. Both exceeded their corresponding uniform-chance levels of
6.25\% and 12.5\%, but remained well below the predefined recovery criteria.
The best within-answer state balanced accuracy was 50.39\%, close to binary
chance, and no location met all qualification criteria. We therefore did not
perform causal patching, examine the two reserved models, or run the provenance
mechanism. This result is limited to one development model, frozen linear
probes, and the predefined token-level sites.

\begin{table}[H]
\caption{Task performance and state localisation in the answer-only
transformer experiment.}
\label{tab:answer-only}
\centering
\small
\begin{tabularx}{\linewidth}{@{}YYrY@{}}
\toprule
Target & Metric & Best result & Criterion \\
\midrule
Model competence
  & held-out task accuracy
  & 100\%
  & $\geq95\%$ \\

Full state
  & 16-way $z_2$ accuracy
  & 23.80\%
  & $\geq70\%$* \\

Answer-relevant component
  & 8-way $z_2\bmod8$ accuracy
  & 45.54\%
  & $\geq90\%$* \\

Within-pair distinction
  & balanced accuracy
  & 50.39\%
  & $\geq75\%$* \\

Qualifying sites
  & frozen 20-site family
  & 0/20
  & $\geq1$ site meeting all criteria \\
\bottomrule
\end{tabularx}
\end{table}

{\footnotesize\noindent
* Qualification also required the registered macro-F1 and minimum-class
thresholds. Full-state qualification additionally required the
answer-relevant and within-answer state criteria.\par}

\subsection{Protected evaluations}
\label{app:protected-evaluations}

A fresh feed-forward model was trained using the fixed 5,000-step procedure,
with a new 128-pair protected set selected and sealed before training. After
passing the public computation, causal, and end-to-end evaluations, the model
was evaluated once on the protected set and passed all 128 pairs.

The engineered transformer was evaluated under the same prospective sequence.
Its protected set was sealed before claim-bearing training, and it was released
only after the model passed the public computation, causal, and end-to-end
tests. The transformer also passed all 128 protected pairs using the unchanged
text signal and detector. Across both protected evaluations, there were no
wrong-state confusions, no detections in the no-signal controls, and no failed
receipt or integrity checks.

In the transformer evaluation, a post-execution evaluator-label mismatch
interrupted the initial adjudication. The primitive evidence had already been
committed before the keys were revealed. Adding the missing positive terminal for the answer-preserving intervention
changed only the evaluator label: no model execution, output, receipt, key,
threshold, or scientific criterion changed. Production and independent
reconstruction then agreed exactly on the unchanged evidence.

\subsection{Threat-model matrix}
\begin{table}[H]
\caption{Trust assumptions and attacks considered in our evaluation.}
\label{tab:threat_model}
\centering
\small
\begin{tabularx}{\linewidth}{@{}lYY@{}}
\toprule
Component & Evaluated assumption & Outside our evaluation \\
\midrule
Model/instrumentation & fixed checkpoint; explicit discrete-state pathway & malicious or uninstrumented computation \\
Receipt system & HMAC integrity; frozen mutation/replay suite & malicious issuer, public verification, key compromise \\
Controller/verifier & correct verification-before-authority ordering & compromised verifier or arbitrary authority selection \\
Surface carrier & fixed grammar and frozen operating point & editing, paraphrase, copy, collusion, open-ended language \\
\bottomrule
\end{tabularx}
\end{table}
\section{Additional construction and evaluation details}

\subsection{Model architecture}
\begin{table}[H]
\caption{Selected Model~1 architecture and trainable parameter count.}
\label{tab:architecture}
\centering
\small
\begin{tabularx}{\linewidth}{@{}lYYr@{}}
\toprule
Component & Input and output & Form & Parameters \\
\midrule
Four embeddings & four 16-way inputs & $16\times16$ each & 1,024 \\
$z_2$ module & three embeddings $\rightarrow16$ categories &
$48\rightarrow128\rightarrow16$, GELU & 8,336 \\
$z_3$ module & hard $z_2$ and $d\rightarrow16$ categories &
$32\rightarrow64\rightarrow16$, GELU & 3,152 \\
Answer head & hard $z_3\rightarrow8$ answers & $16\rightarrow8$ & 136 \\
\midrule
Total & & & 12,648 \\
\bottomrule
\end{tabularx}
\end{table}
The $z_2$ and $z_3$ modules make a single discrete selection using
straight-through argmax; ties are resolved by choosing the lowest-indexed
value. The $z_3$ module receives $z_2$ and $d$, but not $a$, $b$, or $c$,
and the answer head receives only $z_3$. This preserves the mandatory
$z_2\rightarrow z_3\rightarrow y$ pathway described in Section~3.1.

\subsection{Detector calibration}
The detector threshold was fixed using separate calibration data before the
public end-to-end evaluation. Calibration used 512 closures and 16,384
generated outputs. For each closure, we recorded the largest detector score
observed when no state-specific signal was present. The largest of these values,
$T=4.041451884327381$, became the detection threshold, with later outputs
required to exceed it strictly.

We also measured the strongest score assigned to an incorrect state. The
largest calibrated confusion margin was below zero, so the required separation
margin was fixed at $M=0$. A detected state must therefore rank uniquely first,
exceed the threshold $T$, and score strictly above every competing state.

Because each threshold is set from the maximum of 512 calibration examples,
an exchangeable future null example exceeds that maximum with probability at
most $1/513$. Applying this argument separately to no-signal detections and
wrong-state confusions gives a combined bound of
$2/513\approx0.0039$. This is a finite-sample calibration guarantee for the
registered evaluation setting, not a universal false-positive rate.

\subsection{Registered surface vocabulary}
\label{app:surface-vocabulary}

The text generator uses six groups of interchangeable words. Each eligible
position permits the eight alternatives from one group:

\begin{table}[H]
\caption{Permitted alternatives at the variable wording positions.}
\label{tab:surface-vocabulary}
\centering
\small
\begin{tabularx}{\linewidth}{@{}p{0.13\linewidth}p{0.20\linewidth}X@{}}
\toprule
Group & Role & Permitted alternatives \\
\midrule
1 & Opening &
Here, Now, Presently, Directly, Briefly, Simply, Accordingly, Formally \\
2 & Source &
record, calculation, derivation, account, summary, analysis, result, trace \\
3 & Modifier &
plainly, explicitly, carefully, concisely, firmly, notably, precisely,
transparently \\
4 & Reporting verb &
shows, states, gives, reports, records, yields, lists, presents \\
5 & Relation &
that, how, namely, specifically, directly-as, in-form, as-value, with-value \\
6 & Closing &
indeed, therefore, accordingly, thus, thereby, consistently, exactly,
formally-so \\
\bottomrule
\end{tabularx}
\end{table}

The generator may also use one of the following optional phrases, or omit the
phrase entirely:

\[
\begin{split}
\{&\text{Additionally-noted, Separately-recorded, Explicitly-retained,}\\
  &\text{Carefully-preserved, Directly-confirmed, Formally-listed,}\\
  &\text{Briefly-restated, Transparently-given}\}.
\end{split}
\]

For each eligible position and verified state, four of the eight alternatives
are treated as favoured. This state-dependent subset is derived
cryptographically rather than assigning a fixed word to each state. Individual
words therefore do not identify the state; the detector uses the pattern
accumulated across the complete report.

\subsection{Example matched output pair}
\label{app:example}
Section~5 refers to the following matched example from the original public
evaluation. It was selected deterministically as the first example under that
evaluation's registered ordering.
The outputs are reproduced verbatim from the evaluation. The numerical fields
printed under the labels $z_1$, $z_2$, and $z_3$ are fixed report content and
are not the authenticated internal states used by the provenance mechanism.
The authenticated states are established separately by the verified receipts
and are stated in the headings below. The full texts are:
\begin{quote}\small
\textbf{Natural execution (authenticated $z_2=2$, $z_3=6$, $y=6$):}
``Formally the record transparently shows with-value z1 is 7 thus.
Separately-recorded Formally the derivation plainly records as-value z2 is 4
indeed. Directly-confirmed Here the account firmly gives that z3 is 6 exactly.
Transparently-given Accordingly the analysis plainly reports that answer is 6
consistently. Explicitly-retained''

\textbf{Alternative execution (authenticated $z_2'=10$, $z_3'=14$, $y=6$):}
``Formally the record transparently shows with-value z1 is 7 accordingly.
Separately-recorded Formally the calculation explicitly records in-form z2 is
4 thus. Directly-confirmed Now the summary firmly states that z3 is 6 thereby.
Transparently-given Simply the analysis plainly states namely answer is 6
thus. Explicitly-retained''
\end{quote}
The two outputs report the same fixed content, but their authenticated internal
states differ: the natural execution uses $z_2=2\rightarrow z_3=6$, whereas
the alternative execution uses $z_2'=10\rightarrow z_3'=14$. This change in
the authenticated $z_2$ value determines a different state-specific generation
pattern. With the underlying sampling randomness paired across the two
executions, the resulting differences in wording therefore reflect the change
in internal state rather than a change in reported content.

\end{document}